\pdfoutput=1
\documentclass[11pt]{article}

\usepackage[]{ACL2023}

\usepackage{times}
\usepackage{latexsym}
\usepackage{booktabs}
\usepackage{graphicx}
\usepackage{multirow}
\usepackage{times}
\usepackage{amsfonts,amsmath,amssymb,bm}
\usepackage{graphicx}
\usepackage{xcolor}
\usepackage{setspace}
\usepackage{pdflscape}
\usepackage{pdfpages}
\usepackage{enumitem}
\usepackage{rotating}
\usepackage{caption}
\usepackage{subcaption}

\DeclareMathOperator*{\argmax}{arg\,max}

\usepackage{pifont}% http://ctan.org/pkg/pifont
\newcommand{\cmark}{\ding{51}}%
\newcommand{\xmark}{\ding{55}}%

\usepackage[english]{babel}
\usepackage{hyperref}
\addto\captionsenglish{%
}
\addto\extrasenglish{%
}

\usepackage[T1]{fontenc}
\usepackage[utf8]{inputenc}

\usepackage{microtype}

\usepackage{inconsolata}

\title{An Empirical Comparison of LM-based \\ Question and Answer Generation Methods}

\author{Asahi Ushio \and Fernando Alva-Manchego \and Jose Camacho-Collados\\
  Cardiff NLP, School of Computer Science and Informatics, Cardiff University, UK\\
  \texttt{\{UshioA,AlvaManchegoF,CamachoColladosJ\}@cardiff.ac.uk}
  \\}

\begin{document}
\maketitle

% \begin{spacing}{0.98}

\begin{abstract}
Question and answer generation (QAG) consists of generating a set of question-answer pairs given a context (e.g.\ a paragraph). This task has a variety of applications, such as data augmentation for question answering (QA) models, information retrieval and education. In this paper, we establish baselines with three different QAG methodologies that leverage sequence-to-sequence language model (LM) fine-tuning. Experiments show that an end-to-end QAG model, which is computationally light at both training and inference times, is generally robust and outperforms other more convoluted approaches. However, there are differences depending on the underlying generative LM. Finally, our analysis shows that QA models fine-tuned solely on generated question-answer pairs can be competitive when compared to supervised QA models trained on human-labeled data. 

\end{abstract}

\section{Introduction}
% Puri \cite{puri-etal-2020-training}
% \cite{shakeri-etal-2020-end,bartolo-etal-2021-improving}.
% \cite{lewis-etal-2021-paq}

Question and answer generation (QAG) is the task of generating a set of question-answer pairs given an input context such as a document, a paragraph or a sentence. QAG can be applied to develop question answering (QA) models without human supervision \cite{lewis-etal-2019-unsupervised,zhang-bansal-2019-addressing,puri-etal-2020-training} and as a data augmentation mean for QA model understanding \cite{shakeri-etal-2020-end,bartolo-etal-2021-improving}. Moreover, QAG is used as an aid of educational systems \cite{heilman-smith-2010-good,lindberg-etal-2013-generating}, to improve information retrieval models \cite{pyatkin-etal-2021-asking,lewis-etal-2021-paq}, and as a tool for model interpretation \cite{perez-etal-2020-unsupervised,lee-etal-2020-generating}.

QAG stems from question generation (QG) \cite{mitkov-ha-2003-computer,du-etal-2017-learning,zhou2017neural,du-cardie-2018-harvesting}, which consists of generating a question given an answer on the input context. Despite QG being widely studied in the language model era \cite{murakhovska-etal-2022-mixqg,ushio-etal-2022-generative}, QAG is a more complex task, since the answer needs to be generated and not assumed to be part of the input. Therefore, it is unclear what types of QAG models work in practice as no comprehensive comparisons have been established so far. 

\begin{figure}[!t]
    \centering
    \includegraphics[width=\columnwidth]{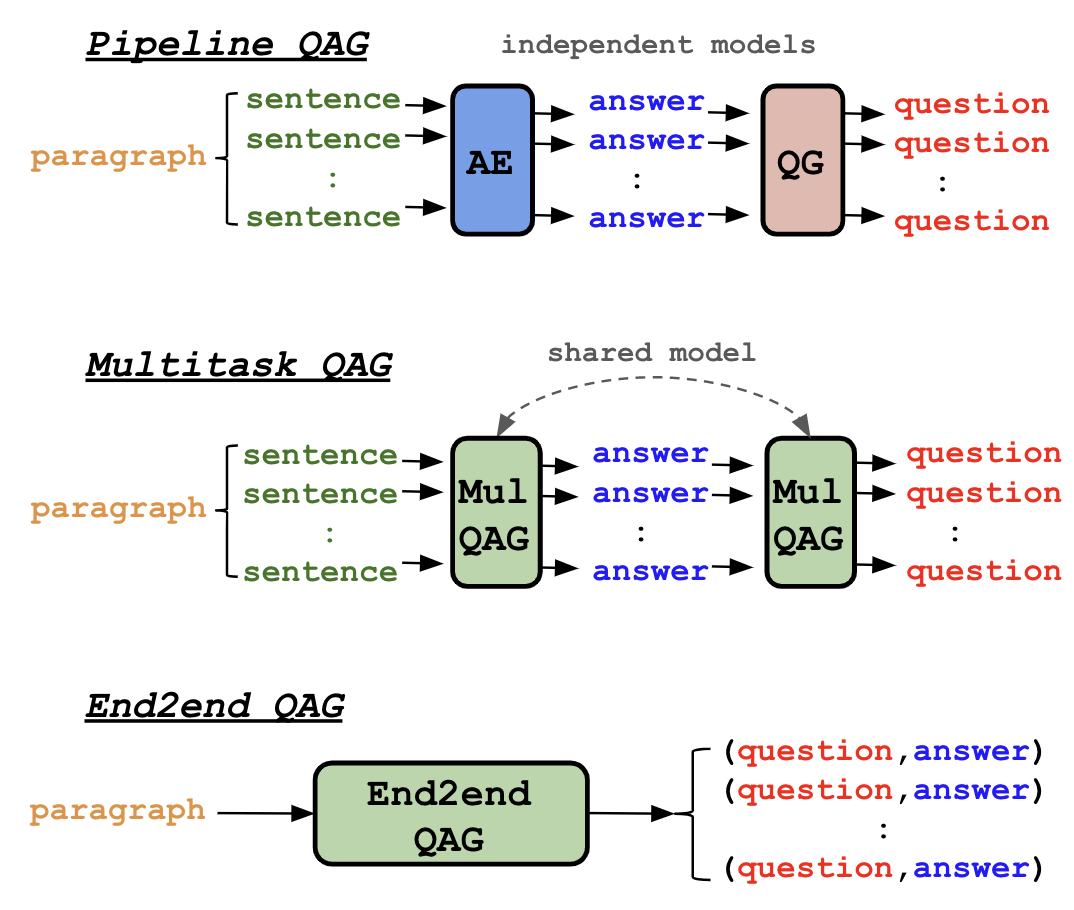}
    \caption{Overview of the considered QAG approaches.}
    \label{fig:overview}
\end{figure}

In this paper, we formalize QAG as a task that generates question-answer pairs given a context, and compare three simple QAG strategies based on fine-tuning encoder-decoder language models (LMs) such as T5 \cite{T5} and BART \cite{lewis-etal-2020-bart}. Our three proposed approaches (illustrated in \autoref{fig:overview}) consist of: (1) pipeline QAG, which decomposes the task into answer extraction and question generation, learning a separate model for each subtask; (2) multitask QAG, which uses a shared single model to train both subtasks instead of independent ones; and (3) end2end QAG, which uses end-to-end sequence-to-sequence learning to generate question-answer pairs directly. Finally, we compare these three approaches on a multi-domain QA-based evaluation, where QA models are trained with the question-answer pairs that each QAG model generates.
All the QAG models are publicly released via HuggingFace \cite{wolf-etal-2020-transformers}\footnote{\url{https://github.com/asahi417/lm-question-generation}}, and available on the online demo\footnote{\url{https://autoqg.net/}}.

\section{Related Work}
There are a few works that leverage pre-trained LMs for QAG. For example, \citet{alberti-etal-2019-synthetic} first fine-tuned BERT \cite{devlin-etal-2019-bert} on answer extraction and QG, and generate question-answer pairs by extracting an answer, on which the associated question is generated. \citet{puri-etal-2020-training} followed a similar idea by fine-tuning an autoregressive LM for QG. In contrast, \citet{shakeri-etal-2020-end} fine-tuned a single LM on answer extraction and QG jointly. \citet{lee-etal-2020-generating} trained an LSTM sequence-to-sequence model from scratch to generate question and answer sequentially. More recently, \citet{bartolo-etal-2021-improving} used a QAG model to generate adversarial examples for QA. Similarly, \citet{lewis-etal-2021-paq} improved on extractive QA by generating millions of question-answer pairs via QAG. In these two last cases, the model to fine-tune was BART \cite{lewis-etal-2020-bart}.

While all these studies use the three methods that we analyse in this paper (i.e. pipeline, multitask and end2end), these are not easily comparable, as there are important differences among them in terms of settings, dataset, input to the LMs, and evaluation metrics.
Moreover, except for \citet{lewis-etal-2021-paq}, none of the proposed QAG models have been made publicly available. Finally, the two most recent studies using BART \cite{bartolo-etal-2021-improving, lewis-etal-2021-paq} have not performed any evaluation on the QAG model, as it is included as a part of a larger pipeline. We summarize the comparison of these prior works and our evaluation at \autoref{tab:comparison-table}.

\section{Question \& Answer Pair Generation} \label{sec:methodology}
Given an input context $c$ (e.g.\ a paragraph), QAG aims to generate natural question-answer pairs $\mathcal{Q}_c$ related to the information in $c$: $\mathcal{Q}_c=\{ (q^1, a^1), (q^2, a^2), \dots \}$. In what follows we describe three different approaches for QAG based on fine-tuning language models.

\begin{table}[!t]
\centering
\scalebox{0.9}{
\begin{tabular}{@{}l@{\hspace{5pt}}c@{\hspace{5pt}}c@{\hspace{5pt}}c@{\hspace{5pt}}c@{\hspace{5pt}}c@{}}
\toprule
                                    & Pipe.                  & Multi.                 & E2E                   & Open           & Eval.                \\ \midrule
\citet{alberti-etal-2019-synthetic}     & \textcolor{green}{\cmark} & \textcolor{red}{\xmark}   & \textcolor{red}{\xmark}   & \textcolor{red}{\xmark}   & \textcolor{green}{\cmark} \\
\citet{puri-etal-2020-training}     & \textcolor{green}{\cmark} & \textcolor{red}{\xmark}   & \textcolor{red}{\xmark}   & \textcolor{red}{\xmark}   & \textcolor{green}{\cmark} \\
\citet{lee-etal-2020-generating}    & \textcolor{red}{\xmark}   & \textcolor{green}{\cmark} & \textcolor{red}{\xmark}   & \textcolor{red}{\xmark}   & \textcolor{green}{\cmark} \\
\citet{shakeri-etal-2020-end}       & \textcolor{green}{\cmark} & \textcolor{red}{\xmark}   & \textcolor{green}{\cmark} & \textcolor{red}{\xmark}   & \textcolor{green}{\cmark} \\
\citet{bartolo-etal-2021-improving} & \textcolor{green}{\cmark} & \textcolor{red}{\xmark}   & \textcolor{red}{\xmark}   & \textcolor{red}{\xmark}   & \textcolor{red}{\xmark}   \\
\citet{lewis-etal-2021-paq}         & \textcolor{green}{\cmark} & \textcolor{red}{\xmark}   & \textcolor{red}{\xmark}   & \textcolor{green}{\cmark}   & \textcolor{red}{\xmark}   \\ \midrule
Ours                                & \textcolor{green}{\cmark} & \textcolor{green}{\cmark} & \textcolor{green}{\cmark} & \textcolor{green}{\cmark} & \textcolor{green}{\cmark} \\
\bottomrule
\end{tabular}
}
\caption{Comparison of our paper and previous studies involving LM-based QAG. The first three columns include the QAG methods used in the corresponding paper: pipeline (Pipe.), multitask (Multi.), and end-to-end (E2E). The fourth column indicates whether QAG models were released open-source (Open). Finally, the last column refers to whether the paper includes QAG evaluation (Eval.).}
\label{tab:comparison-table}
\end{table}

\subsection{Pipeline QAG}\label{pipeline} The QAG task can be decomposed into two simpler subtasks, answer extraction (AE) and QG, where the AE model $P_{{\rm ae}}$ first generates an answer candidate $\tilde{a}$ on a sentence $s$ in context $c$, and then the QG model $P_{{\rm qg}}$ generates a question $\tilde{q}$ that is answerable by answer $\tilde{a}$ given context $c$.
The AE and QG models can be trained independently on any paragraph-level QG datasets that consist of quadruples $(c, s, a, q)$ by maximizing the conditional log-likelihood of:
\begin{align}
    \tilde{a} &= \argmax_{a} P_{{\rm ae}}(a|c, s) \label{eq:answer-extraction} \\
    \tilde{q} &= \argmax_{q} P_{{\rm qg}}(q|c, s, a) \label{eq:question-generation}
\end{align}
where the log-likelihood is factorized into token-level predictions, similar to other sequence-to-sequence learning settings \cite{sutskever2014sequence}. 
In practice, the input to the AE model takes the form of: 
\begin{equation*}
[ c_1, \dots, \texttt{<hl>}, s_1, \dots, s_{|s|}, \texttt{<hl>}, \dots, c_{|c|}  ]
\end{equation*}
where $s_i$ and $c_i$ are the $i-$th token of $s$ and $c$ respectively, $|\cdot|$ represents the number of tokens in a text, and $\texttt{<hl>}$ is the highlighted token to mark the sentence in the context, following the QG formulation of \newcite{chan-fan-2019-recurrent} and \newcite{ushio-etal-2022-generative}.
Likewise, the input to the QG model takes the answer into account by:
\begin{equation*}
[ c_1, \dots, \texttt{<hl>}, a_1, \dots, a_{|a|}, \texttt{<hl>}, \dots, c_{|c|}  ]
\end{equation*}
where $a_i$ is the $i-$th token of $a$.
At inference time, we simply replace the gold answer $a$ of the QG model \eqref{eq:question-generation} by the prediction from the AE model \eqref{eq:answer-extraction}, and run the inference over all the sentences in context $c$ to obtain question-answer pairs.
Consequently, the pipeline approach can generate, at most, as many pairs as sentences in $c$.

\subsection{Multitask QAG}\label{multitask} Instead of training independent models for each subtask, a shared model can be fine-tuned on both AE and QG jointly in a multitask learning manner. To be precise, we mix the training instances for AE and QG altogether, and randomly sample a batch at each iteration of fine-tuning. Each subtask is distinguished by a task prefix added at the beginning of the input text: ``$\texttt{extract answer}$'' (AE) and ``$\texttt{generate question}$'' (QG).

\subsection{End2end QAG}\label{end2end} 
Instead of breaking down QAG into two separate components, we can directly model it by converting the question-answer pairs into a flattened sentence $y$, and fine-tuning a sequence-to-sequence model to generate $y$ from $c$.
Let us define a function that maps $\mathcal{Q}_c$ to a sentence as:
\begin{align}
    \mathcal{T}(\mathcal{Q}_c) &= ``\{t(q^1, a^1)\}\:\texttt{|}\:\{t(q^2, a^2)\}\:\texttt{|}\dots \textrm'\textrm'\label{eq:prompting}\\
    t(q, a)&=``\texttt{question:}\{q\}, \texttt{answer:}\{a\}\textrm'\textrm' \label{eq:prompt-single-pair}
\end{align}
where each pair is textualized with the template \eqref{eq:prompt-single-pair} and joined by a separator $\texttt{|}$.
The end2end QAG model $P_{{\rm qag}}$ is then optimized by maximizing the following conditional log-likelihood:
\begin{align}
    \tilde{y} = \argmax_{y} P_{{\rm qag}}(y|c) \label{eq:qag}
\end{align}

\section{Evaluation}\label{sec:experiments}

\subsection{Experimental Setting} \label{sec:experimental-setting}
\noindent \textbf{Data.} 
QAG models are trained on SQuAD \cite{rajpurkar-etal-2016-squad}.
As their outputs consist of arbitrary questions and answers, reference-based NLG evaluation metrics traditionally used in QG research \cite{papineni-etal-2002-bleu,denkowski-lavie-2014-meteor,lin-2004-rouge,mohammadshahi2022rquge} are unsuitable.
As such, we conduct an extrinsic evaluation by training QA models on the data generated by the QAG models. For this, we rely on SQuADShifts \cite{miller2020effect}, an English reading comprehension dataset in four domains (Amazon/Wikipedia/News/Reddit). For both SQuAD and SQuADShifts, we rely on the train/validation/test splits provided in QG-Bench \cite{ushio-etal-2022-generative}. 

\noindent \textbf{Multi-domain QA Evaluation.}
Given a QAG model to be assessed, we first generate question-answer pairs on each domain of SQuADShifts, and fine-tune DistilBERT \cite{DistilBERT} on the generated pseudo QA pairs, where $F_1$ and exact match on the test set are considered as the target metric. 
This SQuADShifts QA-based evaluation can be used to probe the robustness of the model across domains, as well as for the overall performance by averaging metrics over the domains. Our QA evaluation relies on Tune,\footnote{\url{https://docs.ray.io/en/latest/tune/index.html}} an efficient grid search engine for parameter optimization, to find optimal hyperparameters during QA model fine-tuning.

\noindent \textbf{Base Models.}
For all comparison systems (i.e.\ pipeline, multitask and end2end), we experiment with T5 \cite{T5} and BART \cite{lewis-etal-2020-bart} as base LMs, with the model weights \texttt{t5-\{small,base,large\}} and \texttt{facebook/bart-\{base,large\}} shared on HuggingFace.\footnote{
See \autoref{app:hyper-parameter} for details on the procedure to find optimal hyperparameters during model fine-tuning.} 
Moreover, we report the results of a QG model that takes the gold answers from the provided QA training set as input (QG-only). This is similar to the pipeline method but excluding the AE component.

\subsection{Results} \label{sec:result}

\begin{table}[t]
\centering
\scalebox{0.75}{
\begin{tabular}{@{}l@{\hspace{3pt}}l@{\hspace{5pt}}c@{\hspace{5pt}}c@{\hspace{5pt}}c@{\hspace{5pt}}c@{\hspace{5pt}}c@{}}
\toprule
\multicolumn{2}{@{}l}{Approach}  & Average         & Amazon      & Wiki        & NYT         & Reddit      \\\midrule
\multicolumn{2}{@{}l}{\textit{Gold QA}}                & \textit{53.3/37.3} & \textit{45.9/30.4} & \textit{55.6/38.7} & \textit{61.4/46.9} & \textit{50.1/33.4} \\\midrule
\multirow{4}{*}{\rotatebox{90}{BART\textsubscript{BASE}}}  & QG only             
                                & 49.4/33.9 & 42.3/26.7 & 54.3/37.2 & 59.3/44.8 & 41.9/27.0 \\\cmidrule{2-7}
                   & Pipeline   & 50.0/32.4 & 48.4/29.8 & 49.4/31.1 & 53.0/36.0 & \textbf{49.5}/\textbf{32.7} \\
                   & Multitask  & \textbf{50.8}/\textbf{33.2} & \textbf{49.4}/\textbf{30.5} & \textbf{50.6}/\textbf{32.1} & \textbf{55.0}/\textbf{39.2} & 48.4/31.1\\
                   & End2end    & 34.0/21.4 & 29.3/16.5 & 35.4/23.2 & 44.6/31.1 & 26.6/15.0 \\\midrule
\multirow{4}{*}{\rotatebox{90}{BART\textsubscript{LARGE}}} & QG only            
                                & 49.4/33.8 & 43.3/27.4 & 54.0/36.7 & 59.4/44.6 & 41.1/26.4 \\\cmidrule{2-7}
                    & Pipeline  & 51.7/34.0 & 49.0/30.2 & 52.5/33.7 & 55.3/40.0 & 49.7/32.3 \\
                    & Multitask & \underline{\textbf{54.3}}/\textbf{36.7} & \underline{\textbf{53.6}}/\underline{\textbf{34.3}} & \textbf{54.1}/\textbf{36.4} & \textbf{57.7}/\textbf{41.8} & \underline{\textbf{51.6}}/\underline{\textbf{34.4}}\\
                    & End2end   & 17.5/10.2 & 17.1/9.1  & 18.9/11.3 & 22.3/14.7 & 11.6/5.7  \\\midrule
\multirow{4}{*}{\rotatebox{90}{T5\textsubscript{SMALL}}}   & QG only            
                                & 48.5/33.1 & 43.8/27.7 & 50.5/34.5 & 55.2/41.0 & 44.4/29.1 \\\cmidrule{2-7}
                    & Pipeline  & 45.4/27.4 & 41.6/23.8 & 46.9/27.2 & 48.9/32.1 & \textbf{44.1}/\textbf{26.6} \\
                    & Multitask & 44.0/25.2 & 42.1/22.9 & 44.3/24.1 & 45.6/27.9 & 44.0/25.7 \\
                    & End2end   & \textbf{48.3}/\textbf{31.7} & \textbf{42.3}/\textbf{24.8} & \textbf{54.7}/\textbf{37.2} & \textbf{55.2}/\textbf{40.1} & 41.1/24.8 \\\midrule
\multirow{4}{*}{\rotatebox{90}{T5\textsubscript{BASE}}}    & QG only            
                                & 50.7/34.7 & 43.3/27.4 & 54.4/37.1 & 57.7/43.2 & 47.3/31.1 \\\cmidrule{2-7}
                    & Pipeline  & 51.7/33.6 & \textbf{50.6}/\textbf{31.1} & 52.8/34.0 & 53.8/37.1 & \textbf{49.6}/\textbf{32.4} \\
                    & Multitask & 49.6/30.9 & 48.8/28.9 & 48.1/28.5 & 52.5/35.1 & 49.1/31.2 \\
                    & End2end   & \textbf{51.8}/\textbf{35.4} & 44.9/26.9 & \underline{\textbf{56.9}}/\underline{\textbf{40.1}} & \textbf{59.9}/\textbf{45.3} & 45.3/29.2 \\\midrule
\multirow{4}{*}{\rotatebox{90}{T5\textsubscript{LARGE}}}   & QG only            
                                & 48.9/33.4 & 42.7/26.8 & 53.2/36.2 & 58.5/43.9 & 41.5/26.7 \\\cmidrule{2-7}
                    & Pipeline  & 52.0/33.9 & \textbf{50.9}/\textbf{31.1} & 51.7/32.9 & 55.9/39.5 & \textbf{49.7}/32.0 \\
                    & Multitask & 49.5/30.9 & 46.3/26.0 & 49.6/30.6 & 53.0/35.9 & 49.0/31.2 \\
                    & End2end   & \textbf{53.7}/\underline{\textbf{37.3}} & 49.1/\textbf{31.1} & \textbf{56.1}/\underline{\textbf{40.1}} & \underline{\textbf{60.7}}/\underline{\textbf{45.5}} & 48.8/\textbf{32.5}\\\bottomrule
\end{tabular}
}
\caption{SQuADShifts QA evaluation results ($F_1$/exact match) of different QAG models. As an upperbound, we included the results of the same QA model trained on the gold human-annotated SQuADShifts training set (\textit{Gold QA}). The best score among the QAG approaches within each LM is boldfaced, and the best result in each domain across all models is underlined.}
\label{tab:mqae}
\end{table}

\begin{figure*}[!t]
 \centering
 \includegraphics[width=1.85\columnwidth]{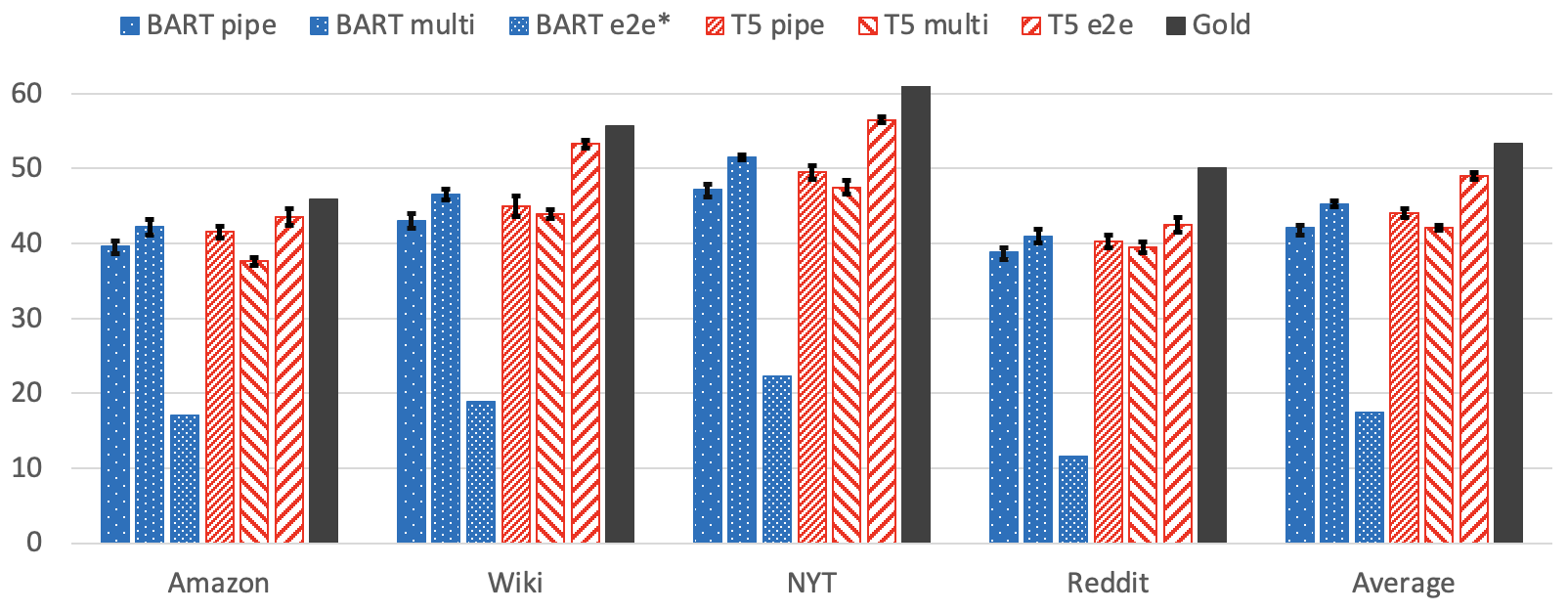}
\caption{Downsampled (equal-sized) SQuADShifts QA evaluation results ($F_1$ score with 95\% confidence interval) for T5\textsubscript{LARGE} multitask/pipeline/end2end and BART\textsubscript{LARGE} pipeline, compared with the original result of each model and the gold QA dataset.}
 \label{fig:downsample}
\end{figure*}

\autoref{tab:mqae} shows the SQuADShifts QA evaluation results for the three approaches considered. Interestingly, the top-2 best models, BART\textsubscript{LARGE} (multitask) and T5\textsubscript{LARGE} (end2end), outperform \textit{Gold QA} (i.e., the model using the human-labeled gold annotations) in two out of four domains, as well as the average in both $F_1$ and exact match. Even smaller models such as T5\textsubscript{SMALL} are competitive with respect to using the gold standard question-answer pairs.

Given the results, it is unclear which approach provides the best performance, as BART\textsubscript{LARGE} (multitask) achieves the best average $F_1$ score (including the best results on Amazon and Reddit domains in both metrics), while T5\textsubscript{LARGE} (end2end) obtains the best average exact match (as well as the best results on Wiki and NYT domains in both metrics). Among the QAG approaches, T5 consistently works better with the end2end QAG, while BART is not well-suited when used end2end. 
A possible explanation is that T5 has observed sentences with structured information due to its multitask pre-training objective, while BART did not have such training instances as it was trained only on a denoising sequence-to-sequence objective.

\subsection{Generation Size Analysis}
\label{sec:analysis}

\begin{table}[t]
\centering
\scalebox{0.80}{
\begin{tabular}{@{}lc@{}}
\toprule
Approach                              & Size (training / validation) \\\midrule
Gold QA                               & 3,141 / 1,571 \\\midrule
% BART\textsubscript{BASE} (pipeline)   & 11,601 / 7,969               \\
% BART\textsubscript{BASE} (multitask)  & 11,392 / 7,825               \\
% BART\textsubscript{BASE} (end2end)    & 963 / 681                    \\
BART\textsubscript{LARGE} (pipeline)  & 11,900 / 8,192               \\
BART\textsubscript{LARGE} (multitask) & 11,752 / 8,103               \\
BART\textsubscript{LARGE} (end2end)   & 2,012 / 1,399                \\
% T5\textsubscript{SMALL} (pipeline)    & 12,017 / 8,277               \\
% T5\textsubscript{SMALL} (multitask)   & 11,989 / 8,242               \\
% T5\textsubscript{SMALL} (end2end)     & 5,969 / 4,130                \\
% T5\textsubscript{BASE} (pipeline)     & 12,211 / 8,404               \\
% T5\textsubscript{BASE} (multitask)    & 12,081 / 8,321               \\
% T5\textsubscript{BASE} (end2end)      & 6,671 / 4,610                \\
T5\textsubscript{LARGE} (pipeline)    & 12,239 / 8,417               \\
T5\textsubscript{LARGE} (multitask)   & 12,148 / 8,357               \\
T5\textsubscript{LARGE} (end2end)     & 6,555 / 4,550                \\
\bottomrule 
\end{tabular}
}
\caption{Average number of question-answer pairs generated for SQuADShifts QA evaluation by each model over all the domains.}
\label{tab:data-size-ave}
\end{table}

In the SQuADShifts QA evaluation, the number of question-answer pairs generated by QAG models is often larger than the human-labelled gold dataset in each domain, as shown in \autoref{tab:data-size-ave}.\footnote{The size of generated question-answer pairs in each domain can be found in \autoref{app:size}.}
Therefore, to fairly compare the quality of generated question-answer pairs, we randomly downsampled the number of the generated question-answer pairs to match the size of the gold dataset. For this analysis we focus on the best-performing T5\textsubscript{LARGE} and BART \textsubscript{LARGE} QAG models\footnote{The end2end BART \textsubscript{LARGE} results match those from \autoref{tab:mqae}, since it had less data than the gold dataset.}, and run the same SQuADShifts QA evaluation with the downsampled pairs.
\autoref{fig:downsample} shows the average of $F_1$ scores over 10 independent trials with different random seeds at downsampling.\footnote{See \autoref{app:ds} for the comparison of exact match.} In this experiment, no model outperforms the gold QA baseline. This indicates that the human-annotated gold dataset is still more informative and data efficient than the generated question-answer pairs. 
Also, since the pipeline/multitask QAG models generate more pairs than the end2end model, downsampling has a larger effect on the pipeline and multitask models than the end2end model. This means that the T5\textsubscript{LARGE} (end2end) model can generate  question-answer pairs of higher quality than those generated by BART\textsubscript{LARGE} (multitask), although they are equally competitive in the main experiment (\autoref{sec:result}).

\subsection{QAG Model Comparison}\label{sec:comparison}

\begin{table}[t]
\centering
\scalebox{0.8}{
\begin{tabular}{@{}l@{\hspace{3pt}}c@{\hspace{5pt}}c@{\hspace{5pt}}c@{}}
\toprule
            & Cost  & Memory& Generated QA   \\\midrule
Pipeline    & 9.2$x$   & 2$x$             &  2.7$x$  \\
Multitask   & 9.2$x$   & $x$             &  2.7$x$  \\
End2end     & $x$     & $x$             &  $x$    \\\bottomrule
\end{tabular}
}
\caption{
Comparison among the three proposed QAG approaches in terms of training cost, memory requirements, and generated question-answer pairs, using end2end as a reference. The comparison is performed for T5\textsubscript{LARGE} with the data used for the main experiments (\autoref{sec:experimental-setting}). Generated QA are averaged across the four SQuADShifts domains.}
\label{tab:model-comparison}
\end{table}
%multitask/pipeline QAG over end2end QAG on the training split from SQuAD.

So far, we have compared the three QAG approaches in terms of performance. However, performance is not the only criterion to consider when choosing a QAG model, since each approach has its own advantages and limitations in terms of computational cost and usability. From the perspective of computational complexity, end2end QAG is faster than the others at both of training and inference, because it can generate a number of question-answer pairs at once in a single paragraph pass. In contrast, both multitask and pipeline need to parse every sentence separately, and a single prediction consists of two generations (i.e. answer extraction and question generation). Essentially, the relative increase of computational cost from end2end QAG to pipeline/multitask QAG can be approximated by the average number of sentences in each paragraph.
In terms of memory requirements, both multitask and end2end QAG rely on a single model, but pipeline QAG consists of two models, requiring twice as much memory storage. Finally, while computational-wise end2end is the lightest model, both pipeline and multitask approaches can generate a larger number of question-answer pairs on average, with the added benefit of being able to run the models on individual sentences. \autoref{tab:model-comparison} shows a practical comparison of the three approaches.
%an example of the relative increase of multitask/pipeline QAG from end2end QAG on the training split of SQuAD \cite{rajpurkar-etal-2016-squad}, which shows clear advantages of the end2end QAG among the three models.
%Apart from complexity, one can specify a sentence $s$ in the context $c$ to generate a question-answer pair with the pipeline/multitask QAG, that results in pipeline/multitask QAG obtaining more pairs than end2end QAG. \autoref{tab:model-comparison} includes an average number of generated question-answer pairs on SQuADShifts\cite{miller2020effect} from all the QAG models fine-tuned on SQuAD in our experiment \autoref{sec:experiments}.

\section{Conclusion}
In this paper, we formalized QAG as a task to generate pairs of questions and answers given an input context, and established baselines with three different QAG approaches. 
To compare them, we conducted a multi-domain QA based evaluation that measures the performance of a QAG model by fine-tuning QA models on the QA training dataset generated by the QAG model. 
Our evaluation shows that end2end QAG models that generate questions and answers simultaneously are generally the most reliable. Nonetheless, establishing a multitask paradigm with separation between answer extraction and question generation can have added benefits, especially when using LMs such as BART. In general, the results are promising, as they show that these artificially-generated QA datasets rival in quality with those annotated by humans, which could save large amount of resources.

\section*{Acknowledgements}

Jose Camacho-Collados is supported by a UKRI Future Leaders Fellowship.

\section*{Limitations}
In this paper, we studied paragraph-level QAG models, which limits their input up to around 500 tokens, and the same approach cannot be easily applied to longer documents. Also, the answer is an entity or a phrase consisting of a few tokens and the question  requires one-hop reasoning, so our models are not able for use in generating longer answers or multi-hop questions. As far as the languages are concerned, the models studies here are English only and to adapt SQuADShifts QA evaluation in other languages, we need QA datasets to train and evaluate the QAG model in those languages. 

The focus on this paper was on evaluating the quality of generated question-answer pairs. As such, we do not attempt to achieve the best QA model possible, but rather use question answering as an extrinsic evaluation. This extrinsic evaluation could be further enhanced with an intrinsic manual evaluation that we did not perform in this paper. Finally, given computational constraints, our QA evaluation is based on a single model only. Again, the goal here was not to achieve the best QA performance, but we acknowledge than using different models could lead to different results.

\section*{Ethics Statement}
Since pre-trained LMs are known to inherit undesirable biases and tend to generate toxic contents in some edge cases \cite{schick-etal-2021-self}, the QAG models we developed in the paper could potentially generate a question or an answer including such texts. Nevertheless, we have done internal validation on the generated question-answer pairs and we have not found such examples in the data analysed in this paper.

\bibliography{anthology,custom}

\begin{thebibliography}{32}
\expandafter\ifx\csname natexlab\endcsname\relax\def\natexlab#1{#1}\fi

\bibitem[{Alberti et~al.(2019)Alberti, Andor, Pitler, Devlin, and
  Collins}]{alberti-etal-2019-synthetic}
Chris Alberti, Daniel Andor, Emily Pitler, Jacob Devlin, and Michael Collins.
  2019.
\newblock \href {https://doi.org/10.18653/v1/P19-1620} {Synthetic {QA} corpora
  generation with roundtrip consistency}.
\newblock In \emph{Proceedings of the 57th Annual Meeting of the Association
  for Computational Linguistics}, pages 6168--6173, Florence, Italy.
  Association for Computational Linguistics.

\bibitem[{Bartolo et~al.(2021)Bartolo, Thrush, Jia, Riedel, Stenetorp, and
  Kiela}]{bartolo-etal-2021-improving}
Max Bartolo, Tristan Thrush, Robin Jia, Sebastian Riedel, Pontus Stenetorp, and
  Douwe Kiela. 2021.
\newblock \href {https://doi.org/10.18653/v1/2021.emnlp-main.696} {Improving
  question answering model robustness with synthetic adversarial data
  generation}.
\newblock In \emph{Proceedings of the 2021 Conference on Empirical Methods in
  Natural Language Processing}, pages 8830--8848, Online and Punta Cana,
  Dominican Republic. Association for Computational Linguistics.

\bibitem[{Chan and Fan(2019)}]{chan-fan-2019-recurrent}
Ying-Hong Chan and Yao-Chung Fan. 2019.
\newblock \href {https://doi.org/10.18653/v1/D19-5821} {A recurrent
  {BERT}-based model for question generation}.
\newblock In \emph{Proceedings of the 2nd Workshop on Machine Reading for
  Question Answering}, pages 154--162, Hong Kong, China. Association for
  Computational Linguistics.

\bibitem[{Denkowski and Lavie(2014)}]{denkowski-lavie-2014-meteor}
Michael Denkowski and Alon Lavie. 2014.
\newblock \href {https://doi.org/10.3115/v1/W14-3348} {Meteor universal:
  Language specific translation evaluation for any target language}.
\newblock In \emph{Proceedings of the Ninth Workshop on Statistical Machine
  Translation}, pages 376--380, Baltimore, Maryland, USA. Association for
  Computational Linguistics.

\bibitem[{Devlin et~al.(2019)Devlin, Chang, Lee, and
  Toutanova}]{devlin-etal-2019-bert}
Jacob Devlin, Ming-Wei Chang, Kenton Lee, and Kristina Toutanova. 2019.
\newblock \href {https://doi.org/10.18653/v1/N19-1423} {{BERT}: Pre-training of
  deep bidirectional transformers for language understanding}.
\newblock In \emph{Proceedings of the 2019 Conference of the North {A}merican
  Chapter of the Association for Computational Linguistics: Human Language
  Technologies, Volume 1 (Long and Short Papers)}, pages 4171--4186,
  Minneapolis, Minnesota. Association for Computational Linguistics.

\bibitem[{Du and Cardie(2018)}]{du-cardie-2018-harvesting}
Xinya Du and Claire Cardie. 2018.
\newblock \href {https://doi.org/10.18653/v1/P18-1177} {Harvesting
  paragraph-level question-answer pairs from {W}ikipedia}.
\newblock In \emph{Proceedings of the 56th Annual Meeting of the Association
  for Computational Linguistics (Volume 1: Long Papers)}, pages 1907--1917,
  Melbourne, Australia. Association for Computational Linguistics.

\bibitem[{Du et~al.(2017)Du, Shao, and Cardie}]{du-etal-2017-learning}
Xinya Du, Junru Shao, and Claire Cardie. 2017.
\newblock \href {https://doi.org/10.18653/v1/P17-1123} {Learning to ask: Neural
  question generation for reading comprehension}.
\newblock In \emph{Proceedings of the 55th Annual Meeting of the Association
  for Computational Linguistics (Volume 1: Long Papers)}, pages 1342--1352,
  Vancouver, Canada. Association for Computational Linguistics.

\bibitem[{Heilman and Smith(2010)}]{heilman-smith-2010-good}
Michael Heilman and Noah~A. Smith. 2010.
\newblock \href {https://aclanthology.org/N10-1086} {Good question! statistical
  ranking for question generation}.
\newblock In \emph{Human Language Technologies: The 2010 Annual Conference of
  the North {A}merican Chapter of the Association for Computational
  Linguistics}, pages 609--617, Los Angeles, California. Association for
  Computational Linguistics.

\bibitem[{Lee et~al.(2020)Lee, Lee, Jeong, Kim, and
  Hwang}]{lee-etal-2020-generating}
Dong~Bok Lee, Seanie Lee, Woo~Tae Jeong, Donghwan Kim, and Sung~Ju Hwang. 2020.
\newblock \href {https://doi.org/10.18653/v1/2020.acl-main.20} {Generating
  diverse and consistent {QA} pairs from contexts with information-maximizing
  hierarchical conditional {VAE}s}.
\newblock In \emph{Proceedings of the 58th Annual Meeting of the Association
  for Computational Linguistics}, pages 208--224, Online. Association for
  Computational Linguistics.

\bibitem[{Lewis et~al.(2020)Lewis, Liu, Goyal, Ghazvininejad, Mohamed, Levy,
  Stoyanov, and Zettlemoyer}]{lewis-etal-2020-bart}
Mike Lewis, Yinhan Liu, Naman Goyal, Marjan Ghazvininejad, Abdelrahman Mohamed,
  Omer Levy, Veselin Stoyanov, and Luke Zettlemoyer. 2020.
\newblock \href {https://doi.org/10.18653/v1/2020.acl-main.703} {{BART}:
  Denoising sequence-to-sequence pre-training for natural language generation,
  translation, and comprehension}.
\newblock In \emph{Proceedings of the 58th Annual Meeting of the Association
  for Computational Linguistics}, pages 7871--7880, Online. Association for
  Computational Linguistics.

\bibitem[{Lewis et~al.(2019)Lewis, Denoyer, and
  Riedel}]{lewis-etal-2019-unsupervised}
Patrick Lewis, Ludovic Denoyer, and Sebastian Riedel. 2019.
\newblock \href {https://doi.org/10.18653/v1/P19-1484} {Unsupervised question
  answering by cloze translation}.
\newblock In \emph{Proceedings of the 57th Annual Meeting of the Association
  for Computational Linguistics}, pages 4896--4910, Florence, Italy.
  Association for Computational Linguistics.

\bibitem[{Lewis et~al.(2021)Lewis, Wu, Liu, Minervini, K{\"u}ttler, Piktus,
  Stenetorp, and Riedel}]{lewis-etal-2021-paq}
Patrick Lewis, Yuxiang Wu, Linqing Liu, Pasquale Minervini, Heinrich
  K{\"u}ttler, Aleksandra Piktus, Pontus Stenetorp, and Sebastian Riedel. 2021.
\newblock \href {https://doi.org/10.1162/tacl_a_00415} {{PAQ}: 65 million
  probably-asked questions and what you can do with them}.
\newblock \emph{Transactions of the Association for Computational Linguistics},
  9:1098--1115.

\bibitem[{Lin(2004)}]{lin-2004-rouge}
Chin-Yew Lin. 2004.
\newblock \href {https://aclanthology.org/W04-1013} {{ROUGE}: A package for
  automatic evaluation of summaries}.
\newblock In \emph{Text Summarization Branches Out}, pages 74--81, Barcelona,
  Spain. Association for Computational Linguistics.

\bibitem[{Lindberg et~al.(2013)Lindberg, Popowich, Nesbit, and
  Winne}]{lindberg-etal-2013-generating}
David Lindberg, Fred Popowich, John Nesbit, and Phil Winne. 2013.
\newblock \href {https://aclanthology.org/W13-2114} {Generating natural
  language questions to support learning on-line}.
\newblock In \emph{Proceedings of the 14th {E}uropean Workshop on Natural
  Language Generation}, pages 105--114, Sofia, Bulgaria. Association for
  Computational Linguistics.

\bibitem[{Miller et~al.(2020)Miller, Krauth, Recht, and
  Schmidt}]{miller2020effect}
John Miller, Karl Krauth, Benjamin Recht, and Ludwig Schmidt. 2020.
\newblock The effect of natural distribution shift on question answering
  models.
\newblock In \emph{International Conference on Machine Learning}, pages
  6905--6916. PMLR.

\bibitem[{Mitkov and Ha(2003)}]{mitkov-ha-2003-computer}
Ruslan Mitkov and Le~An Ha. 2003.
\newblock \href {https://aclanthology.org/W03-0203} {Computer-aided generation
  of multiple-choice tests}.
\newblock In \emph{Proceedings of the {HLT}-{NAACL} 03 Workshop on Building
  Educational Applications Using Natural Language Processing}, pages 17--22.

\bibitem[{Mohammadshahi et~al.(2022)Mohammadshahi, Scialom, Yazdani, Yanki,
  Fan, Henderson, and Saeidi}]{mohammadshahi2022rquge}
Alireza Mohammadshahi, Thomas Scialom, Majid Yazdani, Pouya Yanki, Angela Fan,
  James Henderson, and Marzieh Saeidi. 2022.
\newblock \href {http://arxiv.org/abs/2211.01482} {Rquge: Reference-free metric
  for evaluating question generation by answering the question}.

\bibitem[{Murakhovs{'}ka et~al.(2022)Murakhovs{'}ka, Wu, Laban, Niu, Liu, and
  Xiong}]{murakhovska-etal-2022-mixqg}
Lidiya Murakhovs{'}ka, Chien-Sheng Wu, Philippe Laban, Tong Niu, Wenhao Liu,
  and Caiming Xiong. 2022.
\newblock \href {https://doi.org/10.18653/v1/2022.findings-naacl.111}
  {{M}ix{QG}: Neural question generation with mixed answer types}.
\newblock In \emph{Findings of the Association for Computational Linguistics:
  NAACL 2022}, pages 1486--1497, Seattle, United States. Association for
  Computational Linguistics.

\bibitem[{Papineni et~al.(2002)Papineni, Roukos, Ward, and
  Zhu}]{papineni-etal-2002-bleu}
Kishore Papineni, Salim Roukos, Todd Ward, and Wei-Jing Zhu. 2002.
\newblock \href {https://doi.org/10.3115/1073083.1073135} {{B}leu: a method for
  automatic evaluation of machine translation}.
\newblock In \emph{Proceedings of the 40th Annual Meeting of the Association
  for Computational Linguistics}, pages 311--318, Philadelphia, Pennsylvania,
  USA. Association for Computational Linguistics.

\bibitem[{Perez et~al.(2020)Perez, Lewis, Yih, Cho, and
  Kiela}]{perez-etal-2020-unsupervised}
Ethan Perez, Patrick Lewis, Wen-tau Yih, Kyunghyun Cho, and Douwe Kiela. 2020.
\newblock \href {https://doi.org/10.18653/v1/2020.emnlp-main.713} {Unsupervised
  question decomposition for question answering}.
\newblock In \emph{Proceedings of the 2020 Conference on Empirical Methods in
  Natural Language Processing (EMNLP)}, pages 8864--8880, Online. Association
  for Computational Linguistics.

\bibitem[{Puri et~al.(2020)Puri, Spring, Shoeybi, Patwary, and
  Catanzaro}]{puri-etal-2020-training}
Raul Puri, Ryan Spring, Mohammad Shoeybi, Mostofa Patwary, and Bryan Catanzaro.
  2020.
\newblock \href {https://doi.org/10.18653/v1/2020.emnlp-main.468} {Training
  question answering models from synthetic data}.
\newblock In \emph{Proceedings of the 2020 Conference on Empirical Methods in
  Natural Language Processing (EMNLP)}, pages 5811--5826, Online. Association
  for Computational Linguistics.

\bibitem[{Pyatkin et~al.(2021)Pyatkin, Roit, Michael, Goldberg, Tsarfaty, and
  Dagan}]{pyatkin-etal-2021-asking}
Valentina Pyatkin, Paul Roit, Julian Michael, Yoav Goldberg, Reut Tsarfaty, and
  Ido Dagan. 2021.
\newblock \href {https://doi.org/10.18653/v1/2021.emnlp-main.108} {Asking it
  all: Generating contextualized questions for any semantic role}.
\newblock In \emph{Proceedings of the 2021 Conference on Empirical Methods in
  Natural Language Processing}, pages 1429--1441, Online and Punta Cana,
  Dominican Republic. Association for Computational Linguistics.

\bibitem[{Raffel et~al.(2020)Raffel, Shazeer, Roberts, Lee, Narang, Matena,
  Zhou, Li, and Liu}]{T5}
Colin Raffel, Noam Shazeer, Adam Roberts, Katherine Lee, Sharan Narang, Michael
  Matena, Yanqi Zhou, Wei Li, and Peter~J Liu. 2020.
\newblock Exploring the limits of transfer learning with a unified text-to-text
  transformer.
\newblock \emph{Journal of Machine Learning Research}, 21:1--67.

\bibitem[{Rajpurkar et~al.(2016)Rajpurkar, Zhang, Lopyrev, and
  Liang}]{rajpurkar-etal-2016-squad}
Pranav Rajpurkar, Jian Zhang, Konstantin Lopyrev, and Percy Liang. 2016.
\newblock \href {https://doi.org/10.18653/v1/D16-1264} {{SQ}u{AD}: 100,000+
  questions for machine comprehension of text}.
\newblock In \emph{Proceedings of the 2016 Conference on Empirical Methods in
  Natural Language Processing}, pages 2383--2392, Austin, Texas. Association
  for Computational Linguistics.

\bibitem[{Sanh et~al.(2019)Sanh, Debut, Chaumond, and Wolf}]{DistilBERT}
Victor Sanh, Lysandre Debut, Julien Chaumond, and Thomas Wolf. 2019.
\newblock Distilbert, a distilled version of bert: smaller, faster, cheaper and
  lighter.
\newblock \emph{arXiv preprint arXiv:1910.01108}.

\bibitem[{Schick et~al.(2021)Schick, Udupa, and
  Sch{\"u}tze}]{schick-etal-2021-self}
Timo Schick, Sahana Udupa, and Hinrich Sch{\"u}tze. 2021.
\newblock \href {https://doi.org/10.1162/tacl_a_00434} {Self-diagnosis and
  self-debiasing: A proposal for reducing corpus-based bias in {NLP}}.
\newblock \emph{Transactions of the Association for Computational Linguistics},
  9:1408--1424.

\bibitem[{Shakeri et~al.(2020)Shakeri, Nogueira~dos Santos, Zhu, Ng, Nan, Wang,
  Nallapati, and Xiang}]{shakeri-etal-2020-end}
Siamak Shakeri, Cicero Nogueira~dos Santos, Henghui Zhu, Patrick Ng, Feng Nan,
  Zhiguo Wang, Ramesh Nallapati, and Bing Xiang. 2020.
\newblock \href {https://doi.org/10.18653/v1/2020.emnlp-main.439} {End-to-end
  synthetic data generation for domain adaptation of question answering
  systems}.
\newblock In \emph{Proceedings of the 2020 Conference on Empirical Methods in
  Natural Language Processing (EMNLP)}, pages 5445--5460, Online. Association
  for Computational Linguistics.

\bibitem[{Sutskever et~al.(2014)Sutskever, Vinyals, and
  Le}]{sutskever2014sequence}
Ilya Sutskever, Oriol Vinyals, and Quoc~V Le. 2014.
\newblock Sequence to sequence learning with neural networks.
\newblock \emph{Advances in neural information processing systems}, 27.

\bibitem[{Ushio et~al.(2022)Ushio, Alva-Manchego, and
  Camacho-Collados}]{ushio-etal-2022-generative}
Asahi Ushio, Fernando Alva-Manchego, and Jose Camacho-Collados. 2022.
\newblock {G}enerative language models for paragraph-level question generation.
\newblock In \emph{Proceedings of the 2022 Conference on Empirical Methods in
  Natural Language Processing}, Abu Dhabi, U.A.E. Association for Computational
  Linguistics.

\bibitem[{Wolf et~al.(2020)Wolf, Debut, Sanh, Chaumond, Delangue, Moi, Cistac,
  Rault, Louf, Funtowicz, Davison, Shleifer, von Platen, Ma, Jernite, Plu, Xu,
  Le~Scao, Gugger, Drame, Lhoest, and Rush}]{wolf-etal-2020-transformers}
Thomas Wolf, Lysandre Debut, Victor Sanh, Julien Chaumond, Clement Delangue,
  Anthony Moi, Pierric Cistac, Tim Rault, Remi Louf, Morgan Funtowicz, Joe
  Davison, Sam Shleifer, Patrick von Platen, Clara Ma, Yacine Jernite, Julien
  Plu, Canwen Xu, Teven Le~Scao, Sylvain Gugger, Mariama Drame, Quentin Lhoest,
  and Alexander Rush. 2020.
\newblock \href {https://doi.org/10.18653/v1/2020.emnlp-demos.6} {Transformers:
  State-of-the-art natural language processing}.
\newblock In \emph{Proceedings of the 2020 Conference on Empirical Methods in
  Natural Language Processing: System Demonstrations}, pages 38--45, Online.
  Association for Computational Linguistics.

\bibitem[{Zhang and Bansal(2019)}]{zhang-bansal-2019-addressing}
Shiyue Zhang and Mohit Bansal. 2019.
\newblock \href {https://doi.org/10.18653/v1/D19-1253} {Addressing semantic
  drift in question generation for semi-supervised question answering}.
\newblock In \emph{Proceedings of the 2019 Conference on Empirical Methods in
  Natural Language Processing and the 9th International Joint Conference on
  Natural Language Processing (EMNLP-IJCNLP)}, pages 2495--2509, Hong Kong,
  China. Association for Computational Linguistics.

\bibitem[{Zhou et~al.(2017)Zhou, Yang, Wei, Tan, Bao, and
  Zhou}]{zhou2017neural}
Qingyu Zhou, Nan Yang, Furu Wei, Chuanqi Tan, Hangbo Bao, and Ming Zhou. 2017.
\newblock Neural question generation from text: A preliminary study.
\newblock In \emph{National CCF Conference on Natural Language Processing and
  Chinese Computing}, pages 662--671. Springer.

\end{thebibliography}
\bibliographystyle{acl_natbib}

\clearpage
\appendix

\section{Hyper Parameters}\label{app:hyper-parameter}

\begin{table}[!]
\centering
\scalebox{0.75}{
\begin{tabular}{@{}l@{\hspace{3pt}}l@{\hspace{3pt}}r@{\hspace{8pt}}r@{\hspace{8pt}}r@{\hspace{8pt}}r@{}}
\toprule
Approach      & Model                     & Epoch                & LR      & LS   & Batch \\\midrule
Pipeline (AE) & BART\textsubscript{BASE}  & 4                    & 0.00005 & 0.15 & 64    \\
Pipeline (QG) & BART\textsubscript{BASE}  & 7                    & 0.0001  & 0.15 & 256   \\
Multitask     & BART\textsubscript{BASE}  & 3                    & 0.00005 & 0.15 & 128   \\
End2end       & BART\textsubscript{BASE}  & 2                    & 0.00001 & 0.15 & 128   \\
Pipeline (AE) & BART\textsubscript{LARGE} & 5                    & 0.00005 & 0.15 & 64    \\
Pipeline (QG) & BART\textsubscript{LARGE} & 4                    & 0.00005 & 0.15 & 128   \\
Multitask     & BART\textsubscript{LARGE} &  6      &   0.00001   & 0.15      & 64\\
End2end       & BART\textsubscript{LARGE} & 14                   & 0.00001 & 0.15 & 64    \\
Pipeline (AE) & T5\textsubscript{SMALL}   & 7                    & 0.0001  & 0.15 & 64    \\
Pipeline (QG) & T5\textsubscript{SMALL}   & 9                    & 0.0001  & 0.15 & 64    \\
Multitask     & T5\textsubscript{SMALL}   & 7                    & 0.0001  & 0.15 & 64    \\
End2end       & T5\textsubscript{SMALL}   & 18                   & 0.0001  & 0    & 64    \\
Pipeline (AE) & T5\textsubscript{BASE}    & 8                    & 0.0001  & 0    & 64    \\
Pipeline (QG) & T5\textsubscript{BASE}    & 5                    & 0.0001  & 0.15 & 64    \\
Multitask     & T5\textsubscript{BASE}    & 6                    & 0.0001  & 0.15 & 128   \\
End2end       & T5\textsubscript{BASE}    & 17                   & 0.0001  & 0.15 & 64    \\
Pipeline (AE) & T5\textsubscript{LARGE}   & 9                    & 0.0001  & 0    & 128   \\
Pipeline (QG) & T5\textsubscript{LARGE}   & 6                    & 0.00005 & 0.15 & 64    \\
Multitask     & T5\textsubscript{LARGE}   & 3                    & 0.0001  & 0.15 & 64    \\
End2end       & T5\textsubscript{LARGE}   & 12                   & 0.0001  & 0.15 & 64   \\\bottomrule
\end{tabular}
}
\caption{Optimal hyperparameters for each QAG model.}\label{fig:hyp}
\end{table}

At each QAG model fine-tuning, we search the optimal hyperparameters such as learning rate via \texttt{lmqg}\footnote{\url{https://pypi.org/project/lmqg}}, a hyperparameter search tool for sequence-to-sequence LM fine-tuning, and \autoref{fig:hyp} shows the best hyperparameters.
The maximum input length is fixed as 512, and the maximum output length is 256 for the end2end QAG and 32 for the others.

\section{Size of QA Pairs at SQuADShifts QA evaluation}\label{app:size}

\begin{table}[!t]
\centering
\scalebox{0.75}{
\begin{tabular}{@{}llc@{}}
\toprule
                                   & Approach                         & Size (training / validation) \\\midrule
\multirow{16}{*}{\rotatebox{90}{Amazon}} & Gold QA        & 3,295 / 1,648            \\
                                         & BART\textsubscript{BASE} (pipeline) & 14,824 / 10,273          \\
                                         & BART\textsubscript{LARGE} (pipeline)& 15,204 / 10,569          \\
                                         & T5\textsubscript{SMALL} (pipeline)  & 15,343 / 10,643          \\
                                         & T5\textsubscript{BASE} (pipeline)   & 15,631 / 10,862          \\
                                         & T5\textsubscript{LARGE} (pipeline)  & 15,645 / 10,844          \\
                                         & BART\textsubscript{BASE} (multitask) & 14,517 / 10,065  \\
                                         & BART\textsubscript{LARGE} (multitask)& 15,057 / 10,452   \\
                                         & T5\textsubscript{SMALL} (multitask)  & 15,417 / 10,688          \\
                                         & T5\textsubscript{BASE} (multitask)   & 15,454 / 10,724          \\
                                         & T5\textsubscript{LARGE} (multitask)  & 15,479 / 10,734          \\
                                         & BART\textsubscript{BASE} (end2end) & 990 / 706                \\
                                         & BART\textsubscript{LARGE} (end2end)& 2,045 / 1,408            \\
                                         & T5\textsubscript{SMALL} (end2end)  & 6,419 / 4,470            \\
                                         & T5\textsubscript{BASE} (end2end)   & 7,053 / 4,889            \\
                                         & T5\textsubscript{LARGE} (end2end)  & 7,034 / 4,880            \\\midrule
\multirow{16}{*}{\rotatebox{90}{Wiki}}   & Gold QA       & 2,646 / 1,323            \\
                                         & BART\textsubscript{BASE} (pipeline) & 6,340 / 4,455            \\
                                         & BART\textsubscript{LARGE} (pipeline)& 6,485 / 4,582            \\
                                         & T5\textsubscript{SMALL} (pipeline)  & 6,433 / 4,537            \\
                                         & T5\textsubscript{BASE} (pipeline)   & 6,518 / 4,597            \\
                                         & T5\textsubscript{LARGE} (pipeline)  & 6,518 / 4,596            \\
                                         & BART\textsubscript{BASE} (multitask) & 6,267 / 4,415   \\
                                         & BART\textsubscript{LARGE} (multitask)&  6,450 / 4,547  \\
                                         & T5\textsubscript{SMALL} (multitask)  & 6,377 / 4,504            \\
                                         & T5\textsubscript{BASE} (multitask)   & 6,466 / 4,564            \\
                                         & T5\textsubscript{LARGE} (multitask)  & 6,485 / 4,580            \\
                                         & BART\textsubscript{BASE} (end2end) & 1,137 / 784              \\
                                         & BART\textsubscript{LARGE} (end2end)& 1,718 / 1,214            \\
                                         & T5\textsubscript{SMALL} (end2end)  & 5,050 / 3,513            \\
                                         & T5\textsubscript{BASE} (end2end)   & 5,639 / 3,930            \\
                                         & T5\textsubscript{LARGE} (end2end)  & 5,515 / 3,882            \\\midrule
\multirow{16}{*}{\rotatebox{90}{NYT}}    & Gold QA       & 3,355 / 1,678            \\
                                         & BART\textsubscript{BASE} (pipeline) & 10,033 / 6,913           \\
                                         & BART\textsubscript{LARGE} (pipeline)& 10,339 / 7,141           \\
                                         & T5\textsubscript{SMALL} (pipeline)  & 10,440 / 7,241           \\
                                         & T5\textsubscript{BASE} (pipeline)   & 10,583 / 7,312           \\
                                         & T5\textsubscript{LARGE} (pipeline)  & 10,595 / 7,330           \\
                                         & BART\textsubscript{BASE} (multitask) & 9,857 / 6,781   \\
                                         & BART\textsubscript{LARGE} (multitask)& 10,288 / 7,142   \\
                                         & T5\textsubscript{SMALL} (multitask)  & 10,404 / 7,191           \\
                                         & T5\textsubscript{BASE} (multitask)   & 10,537 / 7,293           \\
                                         & T5\textsubscript{LARGE} (multitask)  & 10,566 / 7,302           \\
                                         & BART\textsubscript{BASE} (end2end) & 1,033 / 756              \\
                                         & BART\textsubscript{LARGE} (end2end)& 2,230 / 1,567            \\
                                         & T5\textsubscript{SMALL} (end2end)  & 6,555 / 4,520            \\
                                         & T5\textsubscript{BASE} (end2end)   & 7,090 / 4,913            \\
                                         & T5\textsubscript{LARGE} (end2end)  & 7,037 / 4,876            \\\midrule
\multirow{16}{*}{\rotatebox{90}{Reddit}} & Gold QA        & 3,268 / 1,634         \\
                                         & BART\textsubscript{BASE} (pipeline) & 15,206 / 10,236       \\
                                         & BART\textsubscript{LARGE} (pipeline)& 15,572 / 10,474       \\
                                         & T5\textsubscript{SMALL} (pipeline)  & 15,853 / 10,688       \\
                                         & T5\textsubscript{BASE} (pipeline)   & 16,112 / 10,844       \\
                                         & T5\textsubscript{LARGE} (pipeline)  & 16,199 / 10,898       \\
                                         & BART\textsubscript{BASE} (multitask) & 14,928 / 10,037   \\
                                         & BART\textsubscript{LARGE} (multitask)& 15,214 / 10,271   \\
                                         & T5\textsubscript{SMALL} (multitask)  & 15,756 / 10,585       \\
                                         & T5\textsubscript{BASE} (multitask)   & 15,866 / 10,704       \\
                                         & T5\textsubscript{LARGE} (multitask)  & 16,063 / 10,813       \\
                                         & BART\textsubscript{BASE} (end2end) & 691 / 477             \\
                                         & BART\textsubscript{LARGE} (end2end)& 2,055 / 1,407        \\
                                         & T5\textsubscript{SMALL} (end2end)  & 5,853 / 4,015         \\
                                         & T5\textsubscript{BASE} (end2end)   & 6,902 / 4,708         \\
                                         & T5\textsubscript{LARGE} (end2end)  & 6,632 / 4,560         \\\bottomrule
\end{tabular}}
\caption{The number of question-answer pairs generated for SQuADShifts QA evaluation in each model.}
\label{tab:data-size}
\end{table}

% \begin{table}[!t]
% \centering
% \scalebox{0.75}{
% \begin{tabular}{lc}
% \toprule
% Approach                              & Size (training / validation) \\\midrule
% BART\textsubscript{BASE} (pipeline)   & 11,601 / 7,969               \\
% BART\textsubscript{BASE} (multitask)  & 11,392 / 7,825               \\
% BART\textsubscript{BASE} (end2end)    & 963 / 681                    \\
% BART\textsubscript{LARGE} (pipeline)  & 11,900 / 8,192               \\
% BART\textsubscript{LARGE} (multitask) & 11,752 / 8,103               \\
% BART\textsubscript{LARGE} (end2end)   & 2,012 / 1,399                \\
% T5\textsubscript{SMALL} (pipeline)    & 12,017 / 8,277               \\
% T5\textsubscript{SMALL} (multitask)   & 11,989 / 8,242               \\
% T5\textsubscript{SMALL} (end2end)     & 5,969 / 4,130              \\
% T5\textsubscript{BASE} (pipeline)     & 12,211 / 8,404               \\
% T5\textsubscript{BASE} (multitask)    & 12,081 / 8,321               \\
% T5\textsubscript{BASE} (end2end)      & 6,671 / 4,610                \\
% T5\textsubscript{LARGE} (pipeline)    & 12,239 / 8,417               \\
% T5\textsubscript{LARGE} (multitask)   & 12,148 / 8,357               \\
% T5\textsubscript{LARGE} (end2end)     & 6,555 / 4,550                \\
% \bottomrule 
% \end{tabular}
% }
% \caption{The average number of question-answer pairs generated for SQuADShifts QA evaluation in each model over all the domains.}
% \label{tab:data-size-ave}
% \end{table}

\autoref{tab:data-size} shows the number of question-answer pairs generated from different QAG models in each domain at SQuADShifts QA evaluation.
The size of the test sets are 4,942 (Amazon), 3,696 (Wiki), 5,032 (NYT), and 4,901 (Reddit).

\section{Additional Results of Downsampled SQuADShifts QA evaluation} \label{app:ds}

\begin{figure*}[!t]
 \centering
 \includegraphics[width=2\columnwidth]{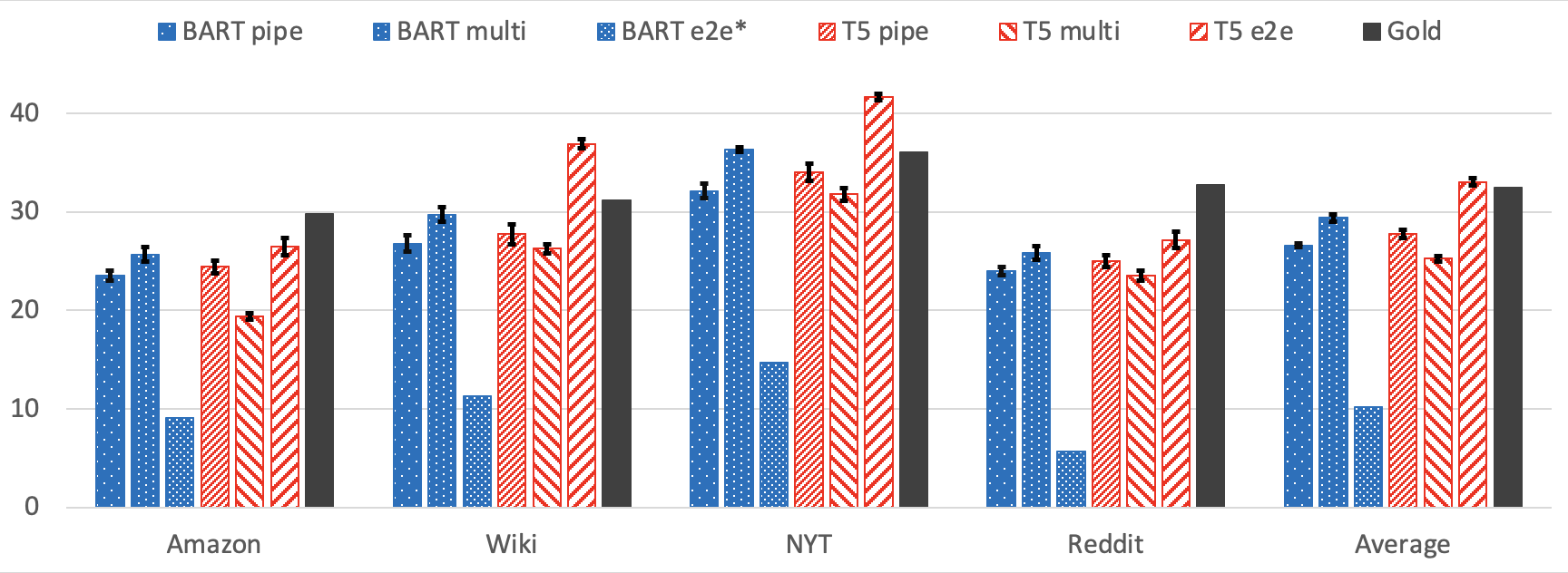}
\caption{Downsampled (equal-sized) SQuADShifts QA evaluation results (exact match with 95\% confidence interval) for T5\textsubscript{LARGE} multitask/pipeline/end2end and BART\textsubscript{LARGE} pipeline, compared with the original result of each model and the gold QA dataset.}
 \label{app:downsample}
\end{figure*}

\autoref{app:downsample} shows the exact match of the downsampled SQuADShifts QA evaluation experiment.

% \end{spacing}

\end{document}